\documentclass[journal,twoside,web]{ieeecolor}
\usepackage{generic}
\usepackage{cite}
\usepackage{amsmath,amssymb,amsfonts}
\usepackage{algorithmic}
\usepackage{graphicx}
\usepackage{algorithm,algorithmic}
\usepackage{hyperref}
\usepackage{array}
\usepackage{booktabs}
\usepackage{adjustbox}
\hypersetup{hidelinks=true}
\usepackage{multirow}
\usepackage{placeins} 
\usepackage{textcomp}
\def\BibTeX{{\rm B\kern-.05em{\sc i\kern-.025em b}\kern-.08em
    T\kern-.1667em\lower.7ex\hbox{E}\kern-.125emX}}
\begin{document}
\title{ERSR: An Ellipse-constrained pseudo-label refinement and symmetric regularization framework for semi-supervised fetal head segmentation in ultrasound images}
\author{Linkuan~Zhou,
 \and
  Zhexin~Chen,
  \and
  Yufei~Shen,
  \and
  Junlin~Xu,
  \and
  Ping~Xuan,
  \and
  Yixin~Zhu,
  \and
  Yuqi~Fang,
  \and
  Cong~Cong,
  \and
  Leyi~Wei,
  \and
  Ran~Su,
  \and
  Jia~Zhou,
  \and
  and~Qiangguo~Jin
  \thanks{Manuscript submitted May 30, 2025. This work was supported by the National Natural Science Foundation of China [Grant No. 62201460, No. 62222311, No. 62322112, and No. 62206197], and the Key Research and Development Program of Shaanxi (Program No. 2025SF-YBXM-424); Linkuan Zhou and Zhexin Chen contributed equally to this work. (Corresponding author: Jia Zhou and Qiangguo Jin)}
  \thanks{Linkuan Zhou, Zhexin Chen, Yufei Shen and Qiangguo Jin are with School of Software, Northwestern Polytechnical University, Shaanxi, China (e-mail: \{zlinkw,chenzhexin,yufeishen\}@mail.nwpu.edu.cn).}
    \thanks{Junlin Xu is with School of Computer Science and Technology, Wuhan University of Science and Technology, Wuhan, China (e-mail: xjl@hnu.edu.cn).}
  \thanks{Ping Xuan is with Department of Computer Science, School of Engineering, Shantou University, Guangdong, China (e-mail: pxuan@stu.edu.cn).}
  \thanks{Yixin Zhu is with Department of Ultrasound, Peking University Shenzhen Hospital, Guangdong, China (e-mail: zyx520@pkuszh.com).}
  \thanks{Yuqi Fang is with School of Intelligence Science and Technology, Nanjing University, Suzhou, China (e-mail: yqfang@nju.edu.cn).}
  \thanks{Cong Cong is with Australian Institute of Health Innovation (AIHI), Macquarie University, Australia (e-mail: thomas.cong@mq.edu.au).}
  \thanks{Leyi Wei is with Faculty of Applied Science, Macao Polytechnic University, Macao SAR, China (e-mail: weileyi@xmu.edu.cn).}
  \thanks{Ran Su is with School of Computer Software, College of Intelligence and Computing, Tianjin University, Tianjin, China (e-mail: ran.su@tju.edu.cn).}
  \thanks{Jia Zhou is with Department of Cardiology, Tianjin Chest Hospital, Tianjin, China (e-mail: zhao\_yifei@tju.edu.cn).}
  \thanks{Qiangguo Jin is also with Yangtze River Delta Research Institute of Northwestern Polytechnical University, Taicang, China (e-mail: qgking@nwpu.edu.cn).}
}

\maketitle

\begin{abstract}
Automated segmentation of the fetal head in ultrasound images is critical for prenatal monitoring. However, achieving robust segmentation remains challenging due to the poor quality of ultrasound images and the lack of annotated data. Semi-supervised methods alleviate the lack of annotated data but struggle with the unique characteristics of fetal head ultrasound images, making it challenging to generate reliable pseudo-labels and enforce effective consistency regularization constraints. To address this issue, we propose a novel semi-supervised framework, ERSR, for fetal head ultrasound segmentation. Our framework consists of the dual-scoring adaptive filtering strategy, the ellipse-constrained pseudo-label refinement, and the symmetry-based multiple consistency regularization. The dual-scoring adaptive filtering strategy uses boundary consistency and contour regularity criteria to evaluate and filter teacher outputs. The ellipse-constrained pseudo-label refinement refines these filtered outputs by fitting least-squares ellipses, which strengthens pixels near the center of the fitted ellipse and suppresses noise simultaneously. The symmetry-based multiple consistency regularization enforces multi-level consistency across perturbed images, symmetric regions, and between original predictions and pseudo-labels, enabling the model to capture robust and stable shape representations. Our method achieves state-of-the-art performance on two benchmarks. On the HC18 dataset, it reaches Dice scores of 92.05\% and 95.36\% with 10\% and 20\% labeled data, respectively. On the PSFH dataset, the scores are 91.68\% and 93.70\% under the same settings.
\end{abstract}


\section{Introduction}
\label{sec:introduction}
\IEEEPARstart{F}{etal} head abnormalities (e.g., microcephaly) are associated with adverse neurodevelopmental outcomes~\cite{kaindl2010many}. Accurate and automated segmentation of the fetal head in ultrasound images is a cornerstone of modern obstetric practice, playing a crucial role in prenatal screening, diagnosis, and monitoring~\cite{torres2022review}.
Precise delineation of the fetal skull enables the quantification of critical biometric parameters, including head circumference (HC), biparietal diameter (BPD), and occipitofrontal diameter (OFD). These measurements are fundamental for assessing gestational age, tracking fetal growth trajectories, detecting potential growth abnormalities such as microcephaly or macrocephaly, and identifying certain congenital malformations~\cite{xodo2023fetal}. In addition, accurate fetal head segmentation is instrumental in guiding prenatal interventions and delivery planning, thereby contributing to improved maternal and fetal outcomes.

The advent of deep learning has revolutionized the field of fetal head segmentation in prenatal ultrasound~\cite{chen2024direction}.
However, the successful application of supervised learning methods is often limited by the need for large-scale, meticulously annotated fetal ultrasound datasets. Acquiring and annotating these datasets is inherently challenging, requiring substantial time and specialized expertise from trained sonographers. Consequently, the limited availability of annotated fetal head ultrasound images hinders the robustness and generalization capability of supervised models. Moreover, accurate fetal head segmentation remains challenging due to blurred or incomplete fetal head boundaries, and interference from black chambers or surrounding maternal tissues~\cite{ansari2024advancements,mushtaq2025comprehensive}.

Recently, semi-supervised learning (SSL) methods have attracted significant interest in the field of medical image segmentation~\cite{yu2019uncertainty,cheplygina2019not,bai2023bidirectional,10246276,su2024mutual,song2024sdcl,bashir2024consistency,WANG2025112767,zeng2025pick}. 
By leveraging both labeled and unlabeled data, SSL offers an effective solution to the challenges posed by the annotation bottleneck.
The SSL techniques can be broadly categorized into consistency regularization~\cite{meanteacher,zhang2017deep,yu2019uncertainty,xu2022all,bai2023bidirectional,bashir2024consistency,lu2024dual}, pseudo-labeling~\cite{huo2021atso,cascante2021curriculum,liu2022semi,10246276,xu2024expectation,zeng2025pick}, and hybrid approaches~\cite{jiang2024intrapartum,su2024mutual,song2024sdcl}. Consistency regularization methods encourage the model to produce stable predictions under different input perturbations. Pseudo-labeling methods generate labels for unlabeled data. Hybrid approaches integrate the strengths of both paradigms by enforcing consistency between student model predictions and pseudo-labels under perturbations, thereby enhancing segmentation performance.

Whereas these SSL techniques have demonstrated promising results in various medical imaging tasks, their direct application to the complex domain of fetal head ultrasound segmentation presents significant challenges.
Firstly, the inherent characteristics of ultrasound images, such as significant noise, acoustic shadowing, speckle artifacts, and ill-defined tissue boundaries, often lead to pseudo-labels with distorted and rough boundaries. These pseudo-labels deviate from the natural elliptical structure of the fetal head, which results in error accumulation during training.
Secondly, current SSL methods, such as Bidirectional Copy-Paste (BCP)~\cite{bai2023bidirectional} and its variants~\cite{su2024mutual,sun2024lcamix}, lack refinement strategies for pseudo-labels that can account for the unique elliptical and symmetrical characteristics of the fetal head, thus limiting their effectiveness in segmentation.
Thirdly, existing methods, in their application of consistency constraints, often overlook potential black chambers that may be present in ultrasound images, as well as the inherent symmetrical structural features of the fetal head. Approaches based on such undiscriminating consistency constraints can disrupt the original anatomical integrity visible in fetal ultrasound images, consequently diminishing the overall effectiveness of consistency-based techniques.

To address the aforementioned limitations, we propose an ellipse-constrained pseudo-label refinement and symmetric regularization framework (ERSR)~\footnote{Code is available at \href{https://github.com/BioMedIA-repo/ERSR}{https://github.com/BioMedIA-repo/ERSR} } for semi-supervised fetal head segmentation in ultrasound images. The ERSR, proposed based on the Mean Teacher (MT) architecture~\cite{meanteacher}, consists of three strategies: the dual-scoring adaptive filtering strategy (DS-AF) employs boundary consistency and contour regularity criteria to evaluate and selectively filter pseudo-labels generated by the teacher model; the ellipse-constrained pseudo-label refinement (EC-PRe) refines the pseudo-labels by fitting least-squares ellipses to the target regions, thereby enhancing pixels near the center of the fitted ellipse and suppressing noise; and the symmetry-based multiple consistency regularization  strategy (SMCR) enforces multi-level consistency across perturbed images, symmetric regions, and between original predictions and pseudo-labels, enabling the model to capture robust and stable shape representations.

The main contributions of this paper are summarized as follows:
\begin{itemize}
\item[$\bullet$] Unlike existing SSL methods that focus on pixel-level consistency, we leverage the elliptical geometric characteristic of the fetal head, an anatomical prior. This issue has been greatly overlooked in prior SSL research, despite its physiological importance in fetal head ultrasound segmentation.
\item[$\bullet$] We propose the dual-scoring adaptive filtering strategy (DS-AF), the ellipse-constrained pseudo-label refinement (EC-PRe), and the symmetry-based multiple consistency regularization  strategy (SMCR). These strategies address the problems of poor pseudo-label quality and error accumulation in the consistency learning process from the perspectives of pseudo-label filtering, pseudo-label refinement, and consistency regularization, respectively.
\item[$\bullet$] The effectiveness of the proposed framework is rigorously validated through extensive experiments on two publicly available fetal head ultrasound datasets, demonstrating state-of-the-art performance compared to existing semi-supervised methods.
\end{itemize}

\section{Related works}
\subsection{Ultrasound fetal head segmentation}
Automated segmentation of the fetal head in ultrasound images has been an active area of research due to its critical role in fetal biometry and prenatal assessment. Early approaches mainly relied on traditional image processing techniques, such as active contour models and statistical shape models~\cite{montagnat2001review}.

The advent of convolutional neural networks (CNNs) has significantly advanced fetal head segmentation in ultrasound imaging~\cite{chen2024direction}. Numerous studies have successfully employed U-Net~\cite{ronneberger2015u} and its variants~\cite{zhou2018unet++, oktay2018attention}, demonstrating substantial improvements over traditional methods.
For instance, Baumgartner~et~al.~\cite{7974824} utilized CNNs to automatically detect 13 standard fetal scan planes and localize fetal structures from 2D ultrasound data in real time, achieving localization using only image-level supervision. Ou~et~al.~\cite{OU2024108501} proposed a real-time segmentation network (RTSeg-Net) for real-time, accurate segmentation of the fetal head and pubic symphysis in intrapartum ultrasound images, specifically designed for deployment on systems with limited hardware capabilities.
Wang~et~al.~\cite{WANG2025112767} combined data augmentation, attention mechanisms, and a progressive training strategy to achieve accurate segmentation of fetal brain structures from a small sample dataset.

Nevertheless, a common limitation of these approaches is their inadequate leverage of unlabeled data~\cite{7974824,OU2024108501}. Moreover, previous methods have often overlooked the geometric characteristics of the fetal head and did not specifically target fetal head features to mitigate the accumulation of pseudo-label errors during training. As a result, they remain heavily reliant on high-precision annotations and large amounts of training data.

\subsection{Semi-supervised learning}
SSL leverages abundant unlabeled data to reduce the reliance on limited labeled samples, significantly alleviating the annotation burden. Mainstream SSL methodologies can be broadly categorized into three groups: consistency regularization, pseudo-labeling, and hybrid approaches.

Consistency regularization is based on the principle that a model's predictions should remain invariant to perturbations applied to its inputs or internal states~\cite{meanteacher}. This is often achieved through data augmentation, with advanced methods such as Mixup~\cite{zhang2017mixup}, Cutout~\cite{devries2017improved}, and CutMix~\cite{yun2019cutmix} being explored to create diverse and challenging perturbations. A canonical example of consistency-based methods is the Mean Teacher framework~\cite{meanteacher}, which enforces prediction consistency between a student model and a teacher model (an exponential moving average of the student's weights).

Pseudo-labeling generates artificial labels for unlabeled data using the model's predictions~\cite{lee2013pseudo}. To mitigate error propagation from incorrect pseudo-labels, strategies like applying a confidence threshold, as seen in FixMatch~\cite{sohn2020fixmatch}, are commonly employed to filter for high-quality labels.

However, directly applying these general-purpose SSL methods to medical image segmentation presents considerable challenges due to the significant domain gap between natural and medical images. Medical images, particularly ultrasound scans, are characterized by low contrast, speckle noise, and ambiguous boundaries, which differ substantially from the rich textures and clear object definitions found in natural images. Consequently, generic data augmentations like CutMix can create anatomically implausible artifacts, potentially corrupting critical diagnostic features rather than promoting robust learning. This necessitates the development of domain-specific adaptations that can effectively handle the unique properties of medical data.

\subsection{Semi-supervised medical image segmentation}
Building on the foundational principles of SSL, numerous methods have been specifically adapted or designed for medical image analysis tasks~\cite{zhang2017deep,yu2019uncertainty,WANG2022102447,bai2023bidirectional,zeng2023pefat,huang2024dual,li2024diversity,zeng2024reciprocal,zeng2025uncertainty,zeng2025pick,zeng2025segment,zeng2025consistency,tang2025swma-unet}. These methods often tailor general SSL strategies to the unique characteristics of medical imaging.

In the context of consistency regularization, methods have been developed to create more semantically meaningful augmentations. For instance, Bidirectional Copy-Paste (BCP)~\cite{bai2023bidirectional} enhances training by copying foreground objects from source images and pasting them onto target images, a technique well-suited for segmentation tasks. Many variants of the Mean Teacher framework have also been proposed to enforce consistency in various medical scenarios~\cite{zhang2017deep,yu2019uncertainty,WANG2022102447,huang2024dual,li2024diversity,zeng2025uncertainty}. For pseudo-labeling, PredICt-and-masK (PICK)~\cite{zeng2025pick} utilizes a multi-decoder architecture to improve the reliability of pseudo-labels for medical images. To address the instability of generated pseudo-labels, Zeng~et~al.~\cite{zeng2023pefat} introduce a loss-based pseudo-label selection method and further utilize unselected low-quality pseudo-labels through feature adversarial training. Hybrid approaches, which combine the strengths of both paradigms, have also demonstrated strong performance across various clinical applications~\cite{cheplygina2019not}. Notable examples include Dual-Student and Teacher Combining CNN and Transformer (DSTCT)~\cite{jiang2024intrapartum}, Mutual~\cite{su2024mutual}, and Students Discrepancy-Informed Correction Learning (SDCL)~\cite{song2024sdcl}, which introduce advanced mechanisms for mutual supervision and bias correction in medical segmentation.

Despite these advancements, many existing semi-supervised medical image segmentation methods still treat medical images as generic data, overlooking crucial anatomical priors that could enhance model performance and stability. For our specific task of fetal head segmentation in ultrasound images, these methods lack explicit constraints on key structural features. This oversight can lead to the accumulation of errors from low-quality pseudo-labels. These limitations motivate our approach, where we aim to integrate task-specific anatomical priors to guide the SSL process more effectively.

\begin{figure*}[ht!]
  \centering
  \includegraphics[width=1\textwidth]{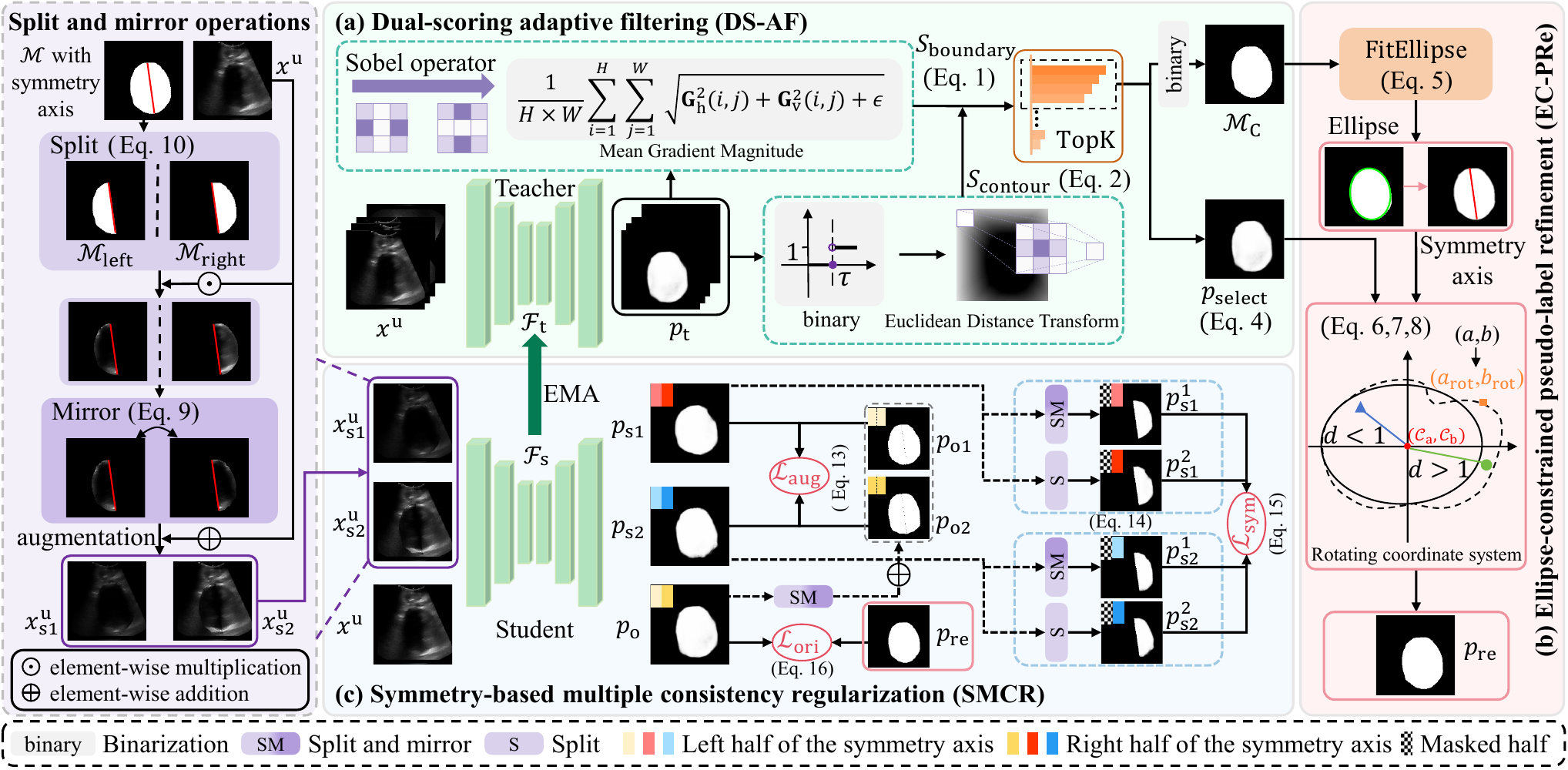}
  \caption{
  Overview of our proposed ellipse-constrained pseudo-label refinement and symmetric regularization (ERSR) framework. In the DS-AF strategy, the $\text{K}$ outputs of the teacher model with the most regular boundaries are selected as pseudo-labels and refined using the EC-PRe strategy. Note that the ``Symmetry axis'' obtained by FitEllipse is used in the ``Split and mirror operations''. Combining the optimized pseudo-labels, we enforce multiple consistency regularization in SMCR. For simplicity, the supervised process of the student model in SMCR is omitted.}
  \label{framework}
\end{figure*}

\section{Method}
\subsection{Overall framework}
Fig.~\ref{framework} shows the proposed ERSR framework for semi-supervised segmentation of fetal head ultrasound images, based on a teacher-student architecture, with pseudo-label filtering, refinement, and consistency regularization. The labeled dataset $\mathcal{X}^{\text{l}} =\left\{(x_{i}, y_{i})\right\}^{N_{\text{l}}}_{i=1}$ and unlabeled dataset $\mathcal{X}^{\text{u}} =\left\{(x_{j})\right\}^{N_{\text{u}}}_{j=1}$ are combined as $\mathcal{X} = \mathcal{X}^{\text{l}} \cup \mathcal{X}^{\text{u}}$, where $N_{\text{l}} << N_{\text{u}}$. For unlabeled data $x^{\text{u}} \in \mathcal{X}^{\text{u}}$, the DS-AF strategy evaluates and filters the output $p_{\text{t}}$ of the teacher model ${\cal{F}}_\text{t}$, resulting in filtered pseudo-labels $p_{\text{selected}}$, which are then refined using the EC-PRe strategy to generate optimized pseudo-labels $p_\text{re}$.
Each $x^{\text{u}}$ is augmented into two symmetric versions, $ x_{\text{s1}}^{\text{u}}$ and $x_{\text{s2}}^{\text{u }}$, via the symmetric random perturbation module. These, along with $x^{\text{u}}$, are fed into the student model ${\cal{F}}_\text{s}$ to predict $p_{\text{o}}$, $p_{\text{s1}}$ and $p_{\text{s2}}$. Prediction $p_{\text{o}}$ is further transformed into two mirrored versions, $p_{\text{o1}}$ and $p_{\text{o2}}$, using predefined symmetry and mirroring operations in symmetric random perturbation. Similarly, predictions $p_{\text{s1}}$ and $p_{\text{s2}}$ are decomposed into $p_{\text{s1}}^{\text{1}}$, $p_{\text{s1}}^{\text{2}}$, $p_{\text{s2}}^{\text{1}}$, and $p_{\text{s2}}^{\text{2}}$. Then SMCR is applied across these predictions, enforcing alignment between $p_{\text{o}}$ and $p_\text{re}$, $p_{\text{s1}}$ and $p_{\text{o1}}$, $p_{\text{s2}}$ and $p_{\text{o2}}$, $p_{\text{s1}}^{\text{1}}$ and $p_{\text{s1}}^{\text{2}}$, as well as $p_{\text{s2}}^{\text{1}}$ and $p_{\text{s2}}^{\text{2}}$. Supervision from labeled data $x^{\text{l}} \in \mathcal{X}^{\text{l}}$ is provided throughout training.

\subsection{Dual-scoring adaptive filtering strategy}
In semi-supervised segmentation, high quality pseudo-labels are essential for accurate learning. Fetal head structures are geometrically continuous and contour coherent. Based on this property, an ideal pseudo-label should similarly ensure boundary consistency and contour geometric regularity. To achieve this, we propose a dual-scoring adaptive filtering strategy, as shown in Fig.~\ref{framework} (a), designed to evaluate predictions from two complementary geometric criteria: boundary consistency and contour regularity, thereby mitigating the error accumulation commonly associated with noisy pseudo-supervision in teacher-student frameworks. Specifically, two distinct geometric scores are computed to assess the quality of the teacher model's predictions, with the highest weighted summation score being selected as reliable pseudo-labels.

\textbf{Boundary consistency score.}
Given the probability map $p_{\text{t}} \in[0,1]^{H\times W}$ from the teacher model ${\cal{F}}_\text{t}$, where $H$ and $W$ are the height and width, respectively. The boundary consistency of $p_{\text{t}}$ is evaluated using the Sobel gradient.
Specifically, we calculate the horizontal and vertical gradients of $p_{\text{t}}$, denoted as $\mathbf{G}_\text{h}$ and $\mathbf{G}_\text{v}$, via the Sobel operator. The boundary consistency score is defined as the mean gradient magnitude:
\begin{equation}S_\mathrm{boundary}=\frac{1}{H\times W}\sum_{i=1}^H\sum_{j=1}^W\sqrt{\mathbf{G}_\text{h}^2(i,j)+\mathbf{G}_\text{v}^2(i,j)+\epsilon},\end{equation}
where $\epsilon$ is a small constant for numerical stability. A lower $S_\mathrm{boundary}$ value indicates a more consistent boundary, and higher values correspond to sharper or noisier boundaries.

\textbf{Contour regularity score.}
To further assess the geometric quantity of $p_{\text{t}}$, we introduce an additional geometric metric based on the Laplacian of the Euclidean Distance Transform ($\text{EDT}$) of its binary mask $\mathcal{M}$. Specifically, the binary mask $\mathcal{M}$ is generated by thresholding $p_{\text{t}}$ at a predefined value $\tau$, such that $\mathcal{M}(i,j) = 1$ if $p_{\text{t}}(i,j) > \tau$, and $\mathcal{M}(i,j) = 0$ otherwise. The $\text{EDT}(\mathcal{M})$ is then computed to produce a distance map that encodes the distance of each foreground pixel to the nearest background pixel. Finally, the distance map is convolved with the $\text{Laplacian}$ operator to obtain a Laplacian value, which reflects local curvature and serves as an indicator of contour regularity. The contour regularity score is calculated as: 
\begin{equation}S_{\mathrm{contour}}=\frac{1}{H\times W}\sum_{i=1}^H\sum_{j=1}^W|\text{EDT}(\mathcal{M}) \ast  \text{Laplacian}(i,j)|,\end{equation}
where $|\cdot |$ denotes the absolute value, and $\ast$ represents the convolution operation between the Euclidean Distance Transform $\text{EDT}(\mathcal{M})$ and the Laplacian kernel. A smaller value of $S_{\mathrm{contour}}$ reflects more continuous structural geometry in the predicted probability map.

The geometric score $S_{\mathrm{score}}$ for each sample is computed by combining the two metrics, $S_\mathrm{boundary}$ and $S_{\mathrm{contour}}$, as follows:
\begin{equation}S_{\mathrm{score}}=1-(\alpha \cdot S_\mathrm{boundary} + (1-\alpha) \cdot S_{\mathrm{contour}}),\end{equation}
where $\alpha \in [0,1]$ is a parameter that balances the boundary consistency score and the contour regularity score.

To select samples dynamically, we introduce a score dictionary $\cal{D}$ that is updated at each training iteration with the latest geometric score ($S_{\mathrm{score}}$) from teacher predictions. This allows us to globally rank all unlabeled samples and select the top $\mathrm{K}$ proportion with the highest scores as reliable supervisory signals:
\begin{equation}
p_{\text{selected}}=\text{TopK}({\cal{D}}(S_{\mathrm{score}}),{\mathrm{K}}),
\end{equation}
where $\text{TopK}(\cdot,\cdot)$ selects the top $\mathrm{K} =  {\lfloor\cal{R}\cdot {\mathit{N}_{\text{u}}}\rfloor}$ samples with the highest $S_{\mathrm{score}}$. The ratio $\cal{R}$ is dynamically adjusted during training via a sigmoid ramp-up function~\cite{laine2016temporal}, prioritizing higher-quality pseudo-labels in the early stages and gradually introducing more samples as training progresses.

In this way, low-quality teacher predictions $p_{\text{t}}$ are filtered out during the early training period, reducing error accumulation and improving overall model stability.

\subsection{Ellipse-constrained pseudo-label refinement}
To address the noisy and irregular boundaries common in pseudo-labels from ultrasound images, we introduce an ellipse-constrained refinement module (Fig.~\ref{framework} (b)). The core rationale is to regularize noisy predictions with a stable geometric prior. To accommodate the natural shape variation where the fetal head only approximates an ellipse, our approach uses ellipse fitting. The goal is not to enforce a perfect ellipse, but rather to capture the dominant global structure while disregarding local boundary imperfections. This refinement provides a more stable and anatomically plausible learning target, steering the model away from fitting to noise.

Given the filtered pseudo-label $p_{\text{selected}}$, the largest connected component $\mathcal{M}_\text{c}$ of its binary mask $\mathcal{M}$ is extracted to focus on the most significant region for refinement. An ellipse is then fitted to this component using a least squares method:
\begin{equation}(\cal{C}_\text{a},\cal{C}_\text{b}) , (\text{axis}_\text{major}, \text{axis}_\text{minor}), \theta = \text{FitEllipse}(\mathcal{M}_\text{c}),\end{equation}
where $\text{FitEllipse}$ denotes the least squares ellipse fitting operation, $(\cal{C}_\text{a},\cal{C}_\text{b})$ represents the center coordinates of the fitted ellipse, $(\text{axis}_\text{major}, \text{axis}_\text{minor})$ denotes the lengths of the major and minor axes, and $\theta$ is the orientation angle.

To effectively utilize the fitted ellipse, spatial coordinates are transformed into a rotating coordinate system aligned with the ellipse's orientation. This transformation is defined as:
\begin{equation}\begin{aligned}
a_{\mathrm{rot}} & =(a-{\cal{C}_\text{a}})\cdot\cos( \frac{\theta \cdot \pi}{180})+(b-{\cal{C}_\text{b}})\cdot\sin(\frac{\theta \cdot \pi}{180}), \\
b_{\mathrm{rot}} & =-(a-{\cal{C}_\text{a}})\cdot\sin(\frac{\theta \cdot \pi}{180})+(b-{\cal{C}_\text{b}})\cdot\cos(\frac{\theta \cdot \pi}{180}),
\end{aligned}\end{equation}
where $(a, b)$ denotes the original pixel coordinates of the point to be rotated. Based on the transformed coordinates, the elliptical distance metric is computed as:
\begin{equation}d=\left(\frac{a_{\mathrm{rot}}}{\text{axis}_\text{major}/2}\right)^2+\left(\frac{b_{\mathrm{rot}}}{\text{axis}_\text{minor}/2}\right)^2.\end{equation}

This metric identifies whether a pixel lies inside ($d \leq 1$) or outside ($d > 1$) the ellipse. Finally, the refined pseudo-label $p_{\text{re}}$ for each pixel is:
\begin{equation}p_\text{re}=
\begin{cases}
\textrm{max}(p_{\text{selected}},\beta+(1-d)^{2}), & d\leq1 ,\\
p_{\text{selected}}\cdot\exp(-(d-1)), & d>1 ,\end{cases}
\label{pre_label}
\end{equation}
where $\beta$ is a small constant that ensures a minimum enhancement for pixels within the ellipse. 

As defined in~\eqref{pre_label}, EC-PRe performs a soft, geometry-guided refinement rather than a hard replacement. It enhances probabilities inside the fitted ellipse and smoothly suppresses them outside, inherently handling deviations from a perfect elliptical shape. This approach nudges the pseudo-label towards a more anatomically plausible form by penalizing distant noise and reinforcing the core region, thus providing a more reliable and structurally consistent supervisory signal.

\subsection{Symmetry-based multiple consistency regularization}
Existing perturbation strategies like BCP~\cite{bai2023bidirectional} are effective for perturbation but ignore the crucial symmetry of the fetal head, creating anatomically implausible training examples. To address this, we introduce the symmetric random perturbation method that uses localized, symmetric changes to create anatomically plausible augmentations. This process yields a more robust supervisory signal for consistency regularization.

\textbf{Symmetric random perturbation.}
To preserve the physiological structure of the original image while introducing meaningful variability, we propose a symmetric random augmentation method that introduces structural perturbations and pixel-level diversity.
Specifically, given the center point coordinates of the fitted ellipse $(\cal{C}_\text{a},\cal{C}_\text{b})$ and its orientation angle $\theta$, the reflected coordinates $(a', b')$ of an original point $(a, b)$ in the image space are computed as:
\begin{equation}\begin{aligned}
(a^\prime,b^\prime)=&2\cdot[((a-{\cal{C}_\text{a}})\cos\theta+
(b-{\cal{C}_\text{b}})\sin\theta)\cdot(\cos\theta,\sin\theta)] \\
&-(a-{\cal{C}_\text{a}}, b-{\cal{C}_\text{b}})+({\cal{C}_\text{a}},{\cal{C}_\text{b}}).\end{aligned}\label{coord}\end{equation}

This transformation generates the symmetric counterpart of each point with respect to the elliptical axis of symmetry. To classify each pixel to a side of the axis, we use the rotated $x$-coordinate $a_{\mathrm{rot}}$ of a point $a$. Then, the binary mask $\mathcal{M}$ derived from the optimized pseudo-label $p_{\text{re}}$ is split into two segments, $\mathcal{M}_{\mathrm{left}}$ and $\mathcal{M}_{\mathrm{right}}$, on each side of the symmetry axis:
\begin{equation}
\mathcal{M}_{\mathrm{left}} = \mathcal{M} \cdot \mathbb{I}(a_{\mathrm{rot}} < 0), \quad \mathcal{M}_{\mathrm{right}} = \mathcal{M} \cdot \mathbb{I}(a_{\mathrm{rot}} \geq 0),
\label{mask12}
\end{equation}
where $\mathbb{I}(\cdot)$ denotes the indicator function, which evaluates to 1 if the condition is true and 0 otherwise.

Using Eq.~(\ref{coord}) and Eq.~(\ref{mask12}), the foreground of the image $x^{\text{u}}$ is extracted and mirrored to generate two symmetrically augmented images, $x_{\text{s1}}^{\text{u}}$ and $x_{\text{s2}}^{\text{u}}$. Each image is composed by combining the background, one half of the foreground, and its mirrored counterpart:
\begin{equation}\begin{aligned}
x_{\text{s1}}^\text{u}(a, b) =& x^\text{u}(a, b) \cdot (1 - \mathcal{M}(a, b)) 
+ x^\text{u}(a, b) \cdot \mathcal{M}_{\text{left}}(a, b)\\&+ x^\text{u}(a', b') \cdot \mathcal{M}_{\text{left}}(a', b'),\\
x_{\text{s2}}^\text{u}(a, b) =& x^\text{u}(a, b) \cdot (1 - \mathcal{M}(a, b)) 
+ x^\text{u}(a, b) \cdot \mathcal{M}_{\text{right}}(a, b) \\&+ x^\text{u}(a', b') \cdot \mathcal{M}_{\text{right}}(a', b'),
\end{aligned}\label{mirroreq}\end{equation}
where $(a, b)$ and $(a', b')$ denote the original and reflected coordinates, respectively. Note that when symmetric random perturbation is enabled, random pixel-level augmentations are applied independently to each half and its mirrored counterpart before recombination, increasing diversity while maintaining structural symmetry.

\textbf{Consistency regularization constraints.}
Combining the original unlabeled image $x^\text{u} \in \mathcal{X}^{\text{u}}$ and its symmetrically augmented counterparts $x_{\text{s1}}^\text{u}$ and  $x_{\text{s2}}^\text{u}$, the student model ${\cal{F}}_\text{s}$ generates three corresponding predictions: $p_{\text{o}}$, $p_{\text{s1}}$, and $p_{\text{s2}}$, respectively. To ensure consistency between original and augmented image predictions, similar to Eq.~(\ref{mirroreq}), the prediction $p_{\text{o}}$ from the original image is symmetrically processed to obtain two mirrored predictions, $p_{\text{o1}}$ and $p_{\text{o2}}$. The value at each point $x=(a,b)$ is determined by its position relative to the symmetry axis, defined by its rotated coordinate $a_{\mathrm{rot}}$:
\begin{equation}
(p_{\text{o1}}(x), p_{\text{o2}}(x)) = 
\begin{cases}
    (p_{\text{o}}(x), p_{\text{o}}(x')), & a_{\text{rot}} < 0 \text{ (left)}, \\
    (p_{\text{o}}(x'), p_{\text{o}}(x)), &  a_{\text{rot}} \geq 0 \text{ (right)},
\end{cases}
\end{equation}
where $x'=(a', b')$ is the symmetric counterpart of $x$. To enforce alignment between the symmetrically transformed predictions and their augmented counterparts, we define the augmentation consistency loss as:
\begin{equation}
    \mathcal{L}_{\text{aug}} = \text{MSE}(p_{\text{o1}}, p_{\text{s}1}) + \text{MSE}(p_{\text{o2}}, p_{\text{s}2}),
\end{equation}
where $\text{MSE}(\cdot, \cdot)$ denotes the mean squared error loss.

To further enhance the model's robustness to different perturbations, consistency constraints are introduced between the predictions of different symmetric augmentations on both sides of the symmetry axis. Specifically, each symmetry-augmented prediction $p_{\text{s}i}$ ($i = 1, 2$) is decomposed into two masked components, denoted as \( p^{m}_{\text{s}i} \) (\( m = 1, 2 \)), corresponding to the left and mirrored right sides, respectively. This process is formulated as:
\begin{equation}
p^{m}_{\text{s}i}(a, b) = 
\begin{cases}
p_{\text{s}i}(a, b) \cdot \mathcal{P}^{\text{s}}_{\text{left}}(a, b), & m = 1, \\
p_{\text{s}i}(a', b') \cdot \mathcal{P}^{\text{s}}_{\text{right}}(a', b'), & m = 2,
\end{cases}
\end{equation}
where $i \in \{1, 2\}$ indexes the two symmetry-augmented predictions $(p_{\text{s}1}, p_{\text{s}2})$, while $m = 1$ and $m = 2$ denote the left and mirrored-right components, respectively. The corresponding symmetry consistency loss is defined as:
\begin{equation}
    \mathcal{L}_{\text{sym}} = \text{MSE}(p_{\text{s}1}^1, p_{\text{s}1}^2) + \text{MSE}(p_{\text{s}2}^1, p_{\text{s}2}^2).
\end{equation}

In addition, to ensure that the original prediction remains faithful to the refined pseudo-label, we apply a direct supervision loss between the original prediction $p_{\text{o}}$ and the optimized pseudo-label $p_{\text{re}}$:
\begin{equation}
    \mathcal{L}_{\text{ori}} = \text{MSE}(p_{\text{o}}, p_{\text{re}}) .
\end{equation}

By this strategy, the model maintains consistency at multiple levels: between different perturbed images, between symmetric regions, and between the original prediction and pseudo-labels. This approach encourages the model to focus on structural patterns, enhancing its ability to preserve the structural information of the fetal head and providing more robust supervision during training.

\subsection{Loss function}
The training strategy is divided into two stages. In the first stage, the model is pre-trained using only the labeled data, and both the teacher and student networks are initialized with the resulting weights. Supervised training continues during this stage to stabilize feature learning.

In the second (semi-supervised) stage, unlabeled data is introduced to further optimize the student network ${\cal{F}}_\text{s}$ by incorporating consistency regularization. During this process, the teacher network ${\cal{F}}_\text{t}$ is updated using the EMA of the student's parameters.

The overall training objective in the semi-supervised stage combines supervised and consistency-based losses:
\begin{equation}
    \mathcal{L}_{\text{total}} =  \mathcal{L}_{\text{sup}}+ \lambda(\mathcal{L}_{\text{ori}} +\mathcal{L}_{\text{aug}} + \mathcal{L}_{\text{sym}}),
\end{equation}
where $\mathcal{L}_{\text{sup}}$ is a weighted sum of the Dice loss and the binary cross-entropy (BCE) loss, which are commonly used in medical image segmentation. The weighting coefficient $\lambda$ is gradually increased from 0 to its maximum value using a sigmoid ramp-up schedule~\cite{laine2016temporal}, allowing the model to focus on reliable supervised signals in the early training phase before progressively incorporating consistency constraints.

\section{Experiments and results}
\subsection{Experiment details}
\textbf{Dataset.} Experiments are conducted on two publicly available datasets: the HC18 dataset and the PSFH dataset. The HC18 dataset, sourced from the 2018 grand challenge~\cite{van2018automated}, contains 1,354 ultrasound images collected from 551 pregnant women. It includes 999 images for training and 355 for testing. After preprocessing, elliptical annotations are filled to create binary segmentation masks for the fetal head.
The PSFH dataset, provided by the MICCAI 2023 grand challenge~\cite{lu2022jnu}, consists of 5,100 ultrasound images extracted from perinatal transperineal ultrasound videos. Of these, 4,000 images are used for training and 1,100 for testing. Each image is paired with a high-quality segmentation mask of the fetal head. Notably, the standard annotations of test sets are not provided. Therefore, only the training sets are used in our experiments, which are further split into training, validation, and test subsets in a 7:1:2 ratio. All images are normalized to zero mean and unit variance, and standard affine transformations such as rotation and scaling are applied for data augmentation.

\textbf{Evaluation metrics.} The performance of the model is evaluated using three standard metrics: Dice similarity coefficient (Dice), 95\% Hausdorff distance ($\text{HD}_{95}$), and average surface distance (ASD).

\textbf{Implementation details.} Our method is implemented in PyTorch and executed on an NVIDIA RTX 3090 GPU. The backbone is set as U-Net~\cite{ronneberger2015u}, optimized using the stochastic gradient descent (SGD) optimizer with a polynomial learning rate schedule. The learning rate is initialized at $1 \times 10^{-2}$ and decayed according to $\left(1-\frac{iter}{total\_{iter}}\right)^{power}$ with $power$ at 0.9. The total number of training iterations is 10,000, and it takes approximately 2 hours to complete these iterations. The batch size is set to 8 for both labeled and unlabeled data. The hyperparameters are empirically set as: $\alpha = 0.5$, $\beta = 0.6$, $\tau = 0.5$, and the initial pseudo-label selection ratio $\cal{R}$ is set to 0.5. All quantitative results are based on the performance of the student model and represent the average over three independent runs.

\subsection{Comparison with state-of-the-art methods}
We compare our method with several state-of-the-art SSL methods, including MT~\cite{meanteacher}, AdvNet~\cite{zhang2017deep}, UAMT~\cite{yu2019uncertainty}, R-Drop~\cite{wu2021r}, BCP~\cite{bai2023bidirectional}, DSTCT~\cite{jiang2024intrapartum}, Mutual~\cite{su2024mutual}, SDCL~\cite{song2024sdcl}, PICK~\cite{zeng2025pick}, and UniMatch V2~\cite{yang2025unimatch}.
Based on their underlying technical approaches, the compared methods are grouped into three categories: consistency regularization~\cite{meanteacher,zhang2017deep,yu2019uncertainty,wu2021r,bai2023bidirectional,yang2025unimatch}, pseudo-labeling~\cite{zeng2025pick}, and hybrid methods~\cite{song2024sdcl,jiang2024intrapartum,su2024mutual}. 

The results in Tables~\ref{HC18} and~\ref{PSFH} demonstrate that our proposed method, ERSR, significantly outperforms the baseline model (U-Net) and achieves superior performance compared to other methods across the Dice, $\text{HD}_{95}$, and ASD metrics under the same labeled data ratio. Notably, on the HC18 dataset with 20\% labeled data, ERSR achieves a Dice of 95.36\%, representing a 3.96\% improvement over the second-best method, PICK~\cite{zeng2025pick}, and falling just 0.95\% short of the supervised upper bound. In contrast, UniMatch V2 underperformed significantly. We attribute this failure to its large ViT encoder's pre-training on natural images. The features learned from this process are ill-suited for the fetal ultrasound domain, which negates potential pre-training advantages.

\begin{table}[h]
\centering
\caption{Quantitative comparison of different methods on the HC18 dataset. Improvements over the {\color{red}\textbf{second-best}} results are highlighted in {\color[rgb]{0,0.5,0}green}}
\label{HC18}
\renewcommand\arraystretch{0.9}
\setlength{\tabcolsep}{0.2mm}
\fontsize{8}{9}\selectfont
\begin{tabular}{l|cc|lll}
\specialrule{1pt}{0pt}{0pt} % 加粗横线，线宽 1pt
                         & \multicolumn{2}{c|}{Scans used}             & \multicolumn{3}{c}{Metrics}                                                                                         \\ \cline{2-6} 
\multirow{-2}{*}{Method} & Labeled             & Unlabeled             & Dice(\%)$\uparrow$                                  & $\text{HD}_{95}$(mm)$\downarrow$                                 & ASD(mm)$\downarrow$                                  \\ \hline
U-Net~\cite{ronneberger2015u}                    & 69(10\%)                 & 0                     & 70.89                                 & 18.15                                & 0.77                                 \\
U-Net~\cite{ronneberger2015u}                    & 139(20\%)                & 0                     & 85.96                                 & 7.04                                 & 0.27                                 \\
U-Net~\cite{ronneberger2015u}                    & 699(All)                   & 0                     & 96.31                                 & 0.59                                 & 0.25                                 \\ \hline
MT~\cite{meanteacher}                       & \multirow{10}{*}{69(10\%)}    &\multirow{10}{*}{630(90\%)}           & 66.44                                 & 23.12                                & 4.50                                 \\
AdvNet~\cite{zhang2017deep}                   & &                       & 68.33                                 & 18.13                                & 2.60                                 \\
UAMT~\cite{yu2019uncertainty}                     & &                       & 72.18                                 & 16.25                                & 1.55                                 \\
R-Drop~\cite{wu2021r}                    & & & 73.57                                 & 15.29                                & 1.34                                 \\
BCP~\cite{bai2023bidirectional}                      && & 86.82                                 & 6.55                                 & {\color{red} \textbf{0.33}} \\
DSTCT~\cite{jiang2024intrapartum}                    && & 68.76                  & 18.14                        & 2.71           \\
Mutual~\cite{su2024mutual}                   &&& 78.33                                 & 11.57                                & 1.18      \\
SDCL~\cite{song2024sdcl}                     &&& 77.74                                 & 12.30                                & 1.27       \\
PICK~\cite{zeng2025pick}                     & && {\color{red} \textbf{88.80}} & {\color{red} \textbf{5.87}} & 0.34    \\
UniMatch V2~\cite{yang2025unimatch}  &&&86.12          &  11.39    &1.05       \\
ERSR                     & && \textbf{92.05}{\color[rgb]{0,0.5,0}\scriptsize$\uparrow$3.25}& \textbf{3.60}{\color[rgb]{0,0.5,0}\scriptsize$\downarrow$2.27}& \textbf{0.17}{\color[rgb]{0,0.5,0}\scriptsize$\downarrow$0.16}\\ \hline
MT~\cite{meanteacher}                       & \multirow{10}{*}{139(20\%)}    &\multirow{10}{*}{560(80\%)}& 85.96                                 & 8.65                                 & 0.62                                 \\
AdvNet~\cite{zhang2017deep}                   & &                     & 73.48            & 17.58             & 1.97                      \\
UAMT~\cite{yu2019uncertainty}                     && & 81.20              & 8.80             & 0.81                    \\
Rdrop~\cite{wu2021r}                    & & & 86.76             & 8.47                & 0.49                \\
BCP~\cite{bai2023bidirectional}                      &&& 91.16          & {\color{red} \textbf{3.67}} & 0.31             \\
DSTCT~\cite{jiang2024intrapartum}                    &&& 78.07                                 & 12.04                                & 0.87                                 \\
Mutual~\cite{su2024mutual}                   &&& 90.70                                 & 3.73                                 & 0.38       \\
SDCL~\cite{song2024sdcl}                     &&& 89.16                                 & 5.41                                 & 0.45        \\
PICK~\cite{zeng2025pick}                     &&& {\color{red} \textbf{91.40}} & 3.82                                 & {\color{red} \textbf{0.28}} \\
UniMatch V2~\cite{yang2025unimatch}  &&&90.92          &  4.56    &0.59       \\
ERSR                     &&& \textbf{95.36}{\color[rgb]{0,0.5,0}\scriptsize$\uparrow$3.96}& \textbf{1.05}{\color[rgb]{0,0.5,0}\scriptsize$\downarrow$2.62}& \textbf{0.16}{\color[rgb]{0,0.5,0}\scriptsize$\downarrow$0.12}\\
\specialrule{1pt}{0pt}{0pt} % 加粗横线，线宽 1pt
\end{tabular}
\end{table}

\begin{figure*}[ht]
  \centering
  \includegraphics[width=0.75\textwidth]{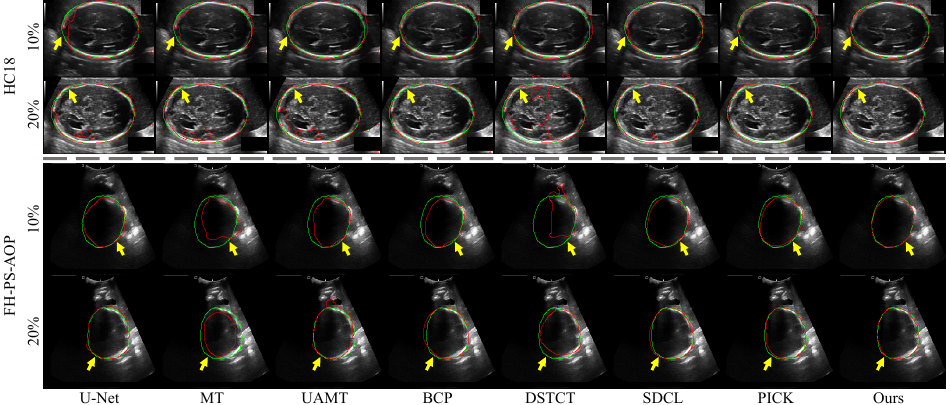}
  \caption{Visual comparison of different methods trained with 10\% and 20\% labeled data. The green line represents the ground truth, while the red line indicates the predicted segmentation. Yellow arrows indicate areas where our method outperforms competing approaches.}
  \label{result}
\end{figure*}

These results indicate that ERSR effectively enables the model to capture structural variations that are consistent with the physiological shape of the fetal head, achieving strong performance on shape-sensitive metrics such as $\text{HD}_{95}$ and ASD, even without explicit boundary supervision. Similarly, as shown in Table~\ref{PSFH}, the effectiveness of ERSR is validated on the PSFH dataset, further demonstrating its generalizability and robustness. Additionally, qualitative results presented in Fig.~\ref{result} show that ERSR produces fewer missegmentations and yields predictions with more regular shapes and smoother boundaries compared to other competing methods.

\begin{table}[ht]
\centering
\caption{Quantitative comparison of different methods on the PSFH dataset. Improvements over the {\color{red}\textbf{second-best}} results are highlighted in {\color[rgb]{0,0.5,0}green}.}
\label{PSFH}
\renewcommand\arraystretch{0.9}
\setlength{\tabcolsep}{0.2mm}
\fontsize{8}{9}\selectfont
\begin{tabular}{l|cc|lll}
\specialrule{1pt}{0pt}{0pt} % 加粗横线，线宽 1pt
                         & \multicolumn{2}{c|}{Scans used}             & \multicolumn{3}{c}{Metrics}                                                                                         \\ \cline{2-6} 
\multirow{-2}{*}{Method} & Labeled             & Unlabeled             & Dice(\%)$\uparrow$                                  & $\text{HD}_{95}$(mm)$\downarrow$                                 & ASD(mm)$\downarrow$                                  \\ \hline
U-Net~\cite{ronneberger2015u}                    &280(10\%)                 & 0                     & 88.82                                 & 4.62                                 & 0.78                                 \\
U-Net~\cite{ronneberger2015u}                    &560(20\%)                 & 0                     & 90.05                                 & 2.91                                 & 0.37                                 \\
U-Net~\cite{ronneberger2015u}                    &2800(All)                   & 0                     & 95.10                                 & 0.70                                 & 0.10                                 \\ \hline
MT~\cite{meanteacher}                       & \multirow{10}{*}{280(10\%)}    &\multirow{10}{*}{2520(90\%)}                   & 82.60                                 & 5.58                                 & 0.92                                 \\
AdvNet~\cite{zhang2017deep}                   &                 &       & 87.87                                 & 4.52                                 & 0.74                                 \\
UAMT~\cite{yu2019uncertainty}                     &                   &   & 82.66                                 & 5.32                                 & 0.89                                 \\
Rdrop~\cite{wu2021r}                    &                   &   & 85.98                                 & 5.08                                 & 0.78     \\
BCP~\cite{bai2023bidirectional}                      &             &    & 88.05                                 & 4.19                                 & 0.69                                 \\
DSTCT~\cite{jiang2024intrapartum}                    &               &    & 71.44                                 & 8.51                                 & 1.91                                 \\
Mutual~\cite{su2024mutual}                   &               &    & 85.73                                 & 5.22            & 0.78       \\
SDCL~\cite{song2024sdcl}                     &        &   & 89.65                                 & 3.97                                 & {\color{red} \textbf{0.58}} \\
PICK~\cite{zeng2025pick}         &           &          & {\color{red} \textbf{89.71}} & {\color{red}\textbf{3.77}} & 0.61   \\
UniMatch V2~\cite{yang2025unimatch}  &&&80.99          &  7.72   &1.37       \\
ERSR                     & && \textbf{91.68}{\color[rgb]{0,0.5,0}\scriptsize$\uparrow$1.97}& \textbf{2.54}{\color[rgb]{0,0.5,0}\scriptsize$\downarrow$1.23}                        & \textbf{0.26}{\color[rgb]{0,0.5,0}\scriptsize$\downarrow$0.32}                        \\ \hline
MT~\cite{meanteacher}                       & \multirow{10}{*}{560(20\%)}    &\multirow{10}{*}{2240(80\%)}                      & 88.47                                 & 5.65                                 & 0.88                                 \\
AdvNet~\cite{zhang2017deep}                   & && 89.54                                 & 3.55                                 & 0.61    \\
UAMT~\cite{yu2019uncertainty}                     & && 89.00                                 & 4.72                                 & 0.96       \\
Rdrop~\cite{wu2021r}                    &&& 88.40                                 & 5.45                                 & 0.89         \\
BCP~\cite{bai2023bidirectional}                      & && 91.28                      & 2.92                & 0.54         \\
DSTCT~\cite{jiang2024intrapartum}                    &&& 89.32                                 & 3.07                                 & 0.72                                 \\
Mutual~\cite{su2024mutual}                   &&& 91.60                                 & 2.85                                 & 0.51         \\
SDCL~\cite{song2024sdcl}                     &&& {\color{red} \textbf{92.96}} & {\color{red} \textbf{1.95}} & 0.36                                 \\
PICK~\cite{zeng2025pick}                     &&& 92.29                      & 2.23                 & {\color{red} \textbf{0.33}} \\
UniMatch V2~\cite{yang2025unimatch}  &&&89.65         &  5.18    &0.94       \\
ERSR                     &&& \textbf{93.70}{\color[rgb]{0,0.5,0}\scriptsize$\uparrow$0.74}& \textbf{1.40}{\color[rgb]{0,0.5,0}\scriptsize$\downarrow$0.55}& \textbf{0.16}{\color[rgb]{0,0.5,0}\scriptsize$\downarrow$0.17}\\
\specialrule{1pt}{0pt}{0pt} % 加粗横线，线宽 1pt
\end{tabular}
\end{table}

\begin{table}[h]
\centering
\caption{Ablation study results on the HC18 dataset. The first row represents the supervised lower bound. Improvements compared with the lower bound in {\color[rgb]{0,0.5,1}blue} are highlighted in {\color[rgb]{0,0.5,0}green}}
\label{ablation_HC18}
\renewcommand\arraystretch{1}
\setlength{\tabcolsep}{0.4mm}
\fontsize{8}{9}\selectfont
 % \begin{adjustbox}{max width=0.5\textwidth}
\begin{tabular}{ccc|cc|ll}
\specialrule{1pt}{0pt}{0pt} % 加粗横线，线宽 1pt
\multicolumn{3}{c|}{Strategy} & \multicolumn{2}{c|}{Scans used}             & \multicolumn{2}{c}{HC18}\\ \cline{1-7}
DS-AF   & EC-PRe  & SMCR  & Labeled              & Unlabeled            & Dice(\%)$\uparrow$& $\text{HD}_{95}$(mm)$\downarrow$\\ \hline
      &         &          &\multirow{7}{*}{69(10\%)}&\multirow{7}{*}{630(90\%)} & {\color[rgb]{0,0.5,1}70.89}& {\color[rgb]{0,0.5,1}18.15}      \\
\checkmark&     &       &            &                   & 87.17      & 7.50       \\
   & \checkmark&      &          &             & 87.54      & 7.33     \\
\checkmark&\checkmark&          &         &            & 87.77      & 6.95   \\
\checkmark&         &\checkmark&           &          & 84.66      & 9.17     \\
    & \checkmark&\checkmark&         &          & 89.94      & 5.61  \\
\checkmark& \checkmark& \checkmark&                      &                      & \textbf{92.05}{\color[rgb]{0,0.5,0}\scriptsize$\uparrow$21.16} &\textbf{3.60}{\color[rgb]{0,0.5,0}\scriptsize$\downarrow$14.55}\\ \hline
&         &          &\multirow{7}{*}{139(20\%)}&\multirow{7}{*}{560(80\%)}& {\color[rgb]{0,0.5,1}85.96}& {\color[rgb]{0,0.5,1}7.04}\\
\checkmark&         &          &        &          & 91.48      & 3.80    \\
     & \checkmark&          &            &          & 91.85      & 3.75    \\
\checkmark& \checkmark&          &         &          & 94.01      & 1.77   \\
\checkmark&         & \checkmark&          &       & 93.74      & 2.14   \\
         & \checkmark& \checkmark&         &          & 95.12      & 1.29  \\
\checkmark& \checkmark& \checkmark&                      &                      & \textbf{95.36}{\color[rgb]{0,0.5,0}\scriptsize$\uparrow$9.4}& \textbf{1.05}{\color[rgb]{0,0.5,0}\scriptsize$\downarrow$5.99}\\
\specialrule{1pt}{0pt}{0pt} % 加粗横线，线宽 1pt
\end{tabular}
% \end{adjustbox}
\end{table}

\begin{table}[ht]
\centering
\caption{Ablation study results on the PSFH dataset. The first row represents the supervised lower bound. Improvements compared with the lower bound in {\color[rgb]{0,0.5,1}blue} are highlighted in {\color[rgb]{0,0.5,0}green}}
\label{ablation}
\renewcommand\arraystretch{1}
\setlength{\tabcolsep}{0.4mm}
\fontsize{8}{9}\selectfont
 % \begin{adjustbox}{max width=0.5\textwidth}
\begin{tabular}{ccc|cc|ll}
\specialrule{1pt}{0pt}{0pt} % 加粗横线，线宽 1pt
\multicolumn{3}{c|}{Strategy} & \multicolumn{2}{c|}{Scans used}           & \multicolumn{2}{c}{PSFH} \\ \cline{1-7}
DS-AF   & EC-PRe  & SMCR  & Labeled              & Unlabeled      & Dice(\%)$\uparrow$& $\text{HD}_{95}$(mm)$\downarrow$\\ \hline
         &         &          &\multirow{7}{*}{280(10\%)}&\multirow{7}{*}{2520(90\%)}      & {\color[rgb]{0,0.5,1}88.82}& {\color[rgb]{0,0.5,1}4.62}\\
 \checkmark&         &          &                      &                        & 90.40      & 3.97      \\
          & \checkmark&          &                      &                       & 90.22      & 4.09      \\
 \checkmark&\checkmark&          &                      &                     & 90.30      & 4.07      \\
 \checkmark&         &\checkmark&                      &                        & 91.19      & 3.09      \\
          & \checkmark&\checkmark&                      &                        & 91.10      & 3.15      \\
\checkmark& \checkmark& \checkmark&                      &                    & \textbf{91.68}{\color[rgb]{0,0.5,0}\scriptsize$\uparrow$2.86}& \textbf{2.54}{\color[rgb]{0,0.5,0}\scriptsize$\downarrow$2.08}\\ \hline
      &       &       &\multirow{7}{*}{560(20\%)}&\multirow{7}{*}{2240(80\%)}& {\color[rgb]{0,0.5,1}90.05}& {\color[rgb]{0,0.5,1}2.91}\\
 \checkmark&         &          &                      &                       & 92.52      & 2.41      \\
          & \checkmark&          &                      &                        & 92.56      & 2.49      \\
 \checkmark& \checkmark&          &                      &                       & 92.47      & 2.43      \\
 \checkmark&         & \checkmark&                      &                       & 93.28      & 1.83      \\
          & \checkmark& \checkmark&                      &                       & 93.46      & 1.67      \\
 \checkmark& \checkmark& \checkmark&                      &                 & \textbf{93.70}{\color[rgb]{0,0.5,0}\scriptsize$\uparrow$3.65}&\textbf{1.40}{\color[rgb]{0,0.5,0}\scriptsize$\downarrow$1.51}\\
\specialrule{1pt}{0pt}{0pt} % 加粗横线，线宽 1pt
\end{tabular}
% \end{adjustbox}
\end{table}

\subsection{Ablation study}
The ablation results in Tables~\ref{ablation_HC18} and~\ref{ablation} reveal not just the individual contribution of each component, but also their sequential and interdependent relationship, functioning as a multi-stage refinement pipeline. Notably, by progressively integrating DS-AF, EC-PRe, and SMCR on the HC18 dataset (10\% labeled), the Dice score improves from a baseline of 70.89\% to 92.05\%. The logic behind this substantial gain is as follows: First, DS-AF removes samples with large errors, providing a clean foundation. Subsequently, EC-PRe acts as a key structural anchor to establish an accurate symmetry axis. Finally, only with a correct geometric prior from EC-PRe can SMCR effectively enforce spatial consistency without introducing errors. This integrated approach, which enforces boundary regularity, geometric rationality, and spatial consistency, establishes a robust refinement process. This structure is essential for preventing error accumulation and demonstrates the indispensable role of each component.

\section{Discussion}
\subsection{Analysis of symmetry axis setting}
In clinical practice, the symmetry axis is a relatively stable structural information of the fetal head, and the ERSR framework implements mirror data augmentation based on the symmetry axis. To determine the most effective axis selection strategy, we systematically evaluate five symmetry axis generation methods: major axis (long), minor axis (short), random rotations around the center point (random), random horizontal (horizontal), and vertical (vertical) axes within mask bounds. The corresponding results are summarized in Table~\ref{axis}.

Notably, results in Table~\ref{axis} show that using anatomically plausible symmetry axes, such as the major (long) and minor (short) axis, yields significantly more accurate results than using randomly defined axes. Among these, the dynamically determined major axis (long) strategy consistently yields the best performance, achieving a Dice score of 92.05\%, which surpasses the second-best approach (vertical axis) by 1.84\%. This validates our choice of using the major axis as a reliable prior for symmetry-based augmentation. Whereas this estimated axis is an effective approximation rather than a guaranteed ground truth, our experiments confirm it is the best available approximation, ensuring robust and effective regularization.

\begin{table}[h]
\centering
\caption{Efficiency analysis of symmetry axis on the HC18 dataset. Improvements over the {\color{red}\textbf{second-best}} results are highlighted in {\color[rgb]{0,0.5,0}green}}
\label{axis}
\renewcommand\arraystretch{0.9}
\setlength{\tabcolsep}{4mm}
\fontsize{8}{9}\selectfont
\begin{tabular}{l|lll}
\specialrule{1pt}{0pt}{0pt} % 加粗横线，线宽 1pt
Strategy   & Dice(\%)$\uparrow$& $\text{HD}_{95}$(mm)$\downarrow$& ASD(mm)$\downarrow$  \\ \hline
short      & 90.19          & 4.48          & 0.19          \\
random     & 83.21          & 10.53         & \color{red} \textbf{0.18}          \\
horizontal & 84.66          & 9.55          & 0.2           \\
vertical   & \color{red}\textbf{90.21}          & \color{red} \textbf{4.27}& 0.19          \\
long (ERSR) & \textbf{92.05}{\color[rgb]{0,0.5,0}\scriptsize$\uparrow$1.84}& \textbf{3.60}{\color[rgb]{0,0.5,0}\scriptsize$\downarrow$0.67}& \textbf{0.17}{\color[rgb]{0,0.5,0}\scriptsize$\downarrow$0.01}\\
\specialrule{1pt}{0pt}{0pt} % 加粗横线，线宽 1pt
\end{tabular}
\end{table}

\subsection{Effectiveness of symmetric random perturbation}
\begin{figure}[ht]
  \centering
  \includegraphics[width=0.4\textwidth]{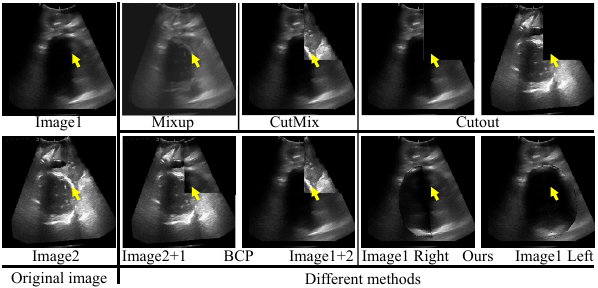}
  \caption{Visual comparison of different image augmentation methods. Yellow arrows highlight the differences in augmentation effects of different data augmentation methods at the same location.}
  \label{mix_vis}
\end{figure}

To further evaluate the effectiveness of our proposed symmetric random perturbation strategy, we compare it with several widely used mixing-based augmentation methods, including Mixup~\cite{zhang2017mixup}, Cutout~\cite{devries2017improved}, CutMix~\cite{yun2019cutmix}, and BCP~\cite{bai2023bidirectional}. A qualitative comparison of these methods on ultrasound images is presented in Fig.~\ref{mix_vis}, while Table~\ref{mix} summarizes their quantitative performance on the HC18 dataset under the 10\% labeled setting. As observed in the visual results, Mixup tends to blur textures and obscure anatomical boundaries, whereas Cutout and CutMix introduce abrupt intensity discontinuities or spatial artifacts that affect structural coherence. BCP reduces these issues by promoting boundary consistency, but its patch-based approach can still disrupt the overall shape of the fetal head. In contrast, our method preserves the natural appearance and the elliptical geometry of the fetal head by enforcing anatomical symmetry during augmentation, improving segmentation performance and retaining anatomical integrity.

\begin{table}[ht]
\centering
\caption{Efficiency analysis of different mixing-based methods on the HC18 dataset. Improvements over the {\color{red}\textbf{second-best}} results are highlighted in {\color[rgb]{0,0.5,0}green}}
\label{mix}
\renewcommand\arraystretch{0.9}
\setlength{\tabcolsep}{4mm}
\fontsize{8}{9}\selectfont
\begin{tabular}{l|lll}
\specialrule{1pt}{0pt}{0pt} % 加粗横线，线宽 1pt
Method & Dice(\%)$\uparrow$& $\text{HD}_{95}$(mm)$\downarrow$& ASD(mm)$\downarrow$  \\ \hline
Mixup~\cite{zhang2017mixup}  & 86.81          & 7.58          & 0.27          \\
Cutout~\cite{devries2017improved} & 84.13          & 9.67          & 0.31          \\
CutMix~\cite{yun2019cutmix} & 86.33          & 7.87          & 0.28          \\
BCP~\cite{bai2023bidirectional}    & \color{red}\textbf{87.97}& \color{red}\textbf{6.34}& \color{red}\textbf{0.26}          \\
ERSR   & \textbf{92.05}{\color[rgb]{0,0.5,0}\scriptsize$\uparrow$4.08} & \textbf{3.60}{\color[rgb]{0,0.5,0}\scriptsize$\downarrow$2.74} & \textbf{0.17}{\color[rgb]{0,0.5,0}\scriptsize$\downarrow$0.09}\\
\specialrule{1pt}{0pt}{0pt} % 加粗横线，线宽 1pt
\end{tabular}
\end{table}

Quantitatively, our method outperforms all competing augmentation strategies. Compared to BCP, it achieves a Dice improvement of 4.08\%, a reduction of 2.74 mm in $\text{HD}_{95}$, and 0.09 mm in ASD. These performance gains stem from two factors: (1) the symmetry-based splitting ensures augmented images remain clinically and anatomically plausible, and (2) the mirror-based augmentation leverages structural information of the fetal head from the same sample without introducing noise from unrelated samples. These improve pseudo-label generation and feature learning, helping the model better capture and generalize anatomical features.

\subsection{Effects of threshold $\tau$ }
Thresholding operations play a critical role in determining the accuracy of both soft probability maps and the resulting hard segmentation masks. As shown in Table~\ref{Threshold}, we systematically evaluate the effect of different binarization thresholds on segmentation performance using the HC18 dataset under the 10\% labeled setting.
The results show that model performance degrades noticeably when the $\tau$ are set too low (e.g., 0.2) or too high (e.g., 0.8), whereas optimal segmentation accuracy is achieved at a threshold of 0.5. Specifically, this threshold yields the highest Dice score (92.05\%), the lowest $\text{HD}_{95}$ (3.60 mm), and the lowest ASD (0.17 mm), surpassing the second-best configuration by a substantial margin. This suggests that thresholds closer to the extremes (0 or 1) introduce bias toward either over-segmentation or under-segmentation, leading to degraded geometric and boundary accuracy, as evidenced by the increase in $\text{HD}_{95}$ and ASD. Therefore, we set the threshold to 0.5 for optimal performance.

\begin{table}[ht]
\centering
\caption{Efficiency analysis of threshold on the HC18 dataset. Improvements over the {\color{red}\textbf{second-best}} results are highlighted in {\color[rgb]{0,0.5,0}green}}
\label{Threshold}
\renewcommand\arraystretch{0.9}
\setlength{\tabcolsep}{4mm}
\fontsize{8}{9}\selectfont
\begin{tabular}{l|lll}
\specialrule{1pt}{0pt}{0pt} % 加粗横线，线宽 1pt
Threshold& Dice(\%)$\uparrow$& $\text{HD}_{95}$(mm)$\downarrow$& ASD(mm)$\downarrow$  \\ \hline
0.2
& 85.48& 8.71& 0.22
\\
0.4
& \color{red}\textbf{90.11}& \color{red}\textbf{5.14}& 0.21
\\
0.5 (ERSR)
& \textbf{92.05}{\color[rgb]{0,0.5,0}\scriptsize$\uparrow$1.94}& \textbf{3.60}{\color[rgb]{0,0.5,0}\scriptsize$\downarrow$1.54}& \textbf{0.17}{\color[rgb]{0,0.5,0}\scriptsize$\downarrow$0.02}
\\
0.6
& 89.59& 5.37& \color{red}\textbf{0.19}
\\
0.8
& 87.35& 7.32& 0.22
\\
\specialrule{1pt}{0pt}{0pt} % 加粗横线，线宽 1pt
\end{tabular}
\end{table}

\subsection{Parameter settings}
\begin{figure}[t]
  \centering
  \includegraphics[width=0.35\textwidth]{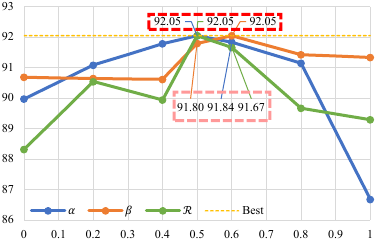}
  \caption{Comparison of different hyperparameter settings on the HC18 dataset. The Y-axis represents the Dice score (\%), and the X-axis denotes the value of each hyperparameter. The optimal values are highlighted in the {\color{red}red} dashed box, whereas the suboptimal values are indicated by the {\color[rgb]{1,0.41,0.71}pink} dashed box.}
  \label{para}
\end{figure}

The hyperparameters $\alpha$, $\beta$, and $\mathcal{R}$ in the ERSR framework control different aspects of the training process: $\alpha$ balances the geometric scoring mechanism, $\beta$ defines the lower bound for pseudo-label enhancement, and $\mathcal{R}$ adjusts the filtering ratio in the Top-K pseudo-label selection. To determine the optimal values, each hyperparameter is systematically varied from 0 to 1 in increments of 0.2, with an additional evaluation at 0.5 to provide finer granularity. The resulting segmentation performance, measured by the Dice score on the HC18 dataset with 10\% labeled data, is shown in Fig.~\ref{para}.
As highlighted by the red dashed box, the best performance (92.05\%) is achieved when $\alpha$ and $\mathcal{R}$ are set to 0.5, and the performance reaches its peak when $\beta$ is 0.6.

\subsection{Analysis with other SSL ultrasound methods}
Recent semi-supervised methods, exemplified by work from the 2025 ISBI Challenge on Fetal Ultrasound Grand Challenges, primarily mitigate pseudo-label noise through two main strategies: enhancing model diversity via multi-teacher frameworks (e.g., COMT~\cite{chen2025comt} and AD-MT~\cite{yu2025enhancing}) or employing advanced data augmentations for consistency training (e.g., HACL~\cite{xiao2025hierarchical} and Semi-CervixSeg~\cite{jiang2025semi}). These approaches enforce consistency under various model perspectives or stochastic perturbations. 

In contrast, our ERSR method adopts a knowledge-driven strategy, integrating the fetal head's inherent symmetry as a strong anatomical prior. Unlike methods that rely on stochastic augmentations for consistency training, ERSR's regularization is based on a deterministic, anatomically valid property, providing a more direct and reliable supervisory signal against pseudo-label noise. Thus, whereas other SSL methods offer broad applicability, ERSR's task-specificity serves as a complementary strength for symmetrical structures, suggesting promising avenues for hybrid approaches that combine both paradigms.

\subsection{Generalizability of ERSR for Elliptical Structures}
The applicability of ERSR's ellipse-constrained refinement to other anatomical structures is a key consideration. To empirically test its generalizability, we evaluate ERSR on the pubic symphysis segmentation task from the PSFHS challenge~\cite{bai2025psfhs}. The method achieves promising results on 20\% labeled pubic symphysis segmentation (Dice: 84.78\%, HD95: 5.3 mm), which confirms the mechanism's generalizability to other structures with a similar elliptical geometry.

We further evaluate the potential applicability of ERSR to fetal abdominal segmentation (ACOUSLIC-AI Challenge~\cite{sappia2025acouslic}). The abdomen is often approximated as an ellipse for biometric measurements, which suggests ERSR could refine segmentations and improve measurement consistency. However, a significant challenge arises from the abdomen's nature as a highly deformable soft-tissue structure, unlike the more rigid fetal skull or pubic symphysis. A direct application of ERSR's constraint risks over-regularizing correct but non-elliptical boundaries. Therefore, adapting ERSR for the abdomen would likely require a more flexible constraint method, which we identify as an important direction for future research.

\subsection{Computational Efficiency and Clinical Practicality}
Computational efficiency is critical for clinical deployment. As shown in Table~\ref{numpara}, our ERSR framework is designed to be exceptionally lightweight. A key advantage of our approach is that our proposed modules (DS-AF, EC-PRe, SMCR) are implemented purely as a training strategy and add no architectural overhead during inference. Consequently, ERSR achieves its state-of-the-art accuracy while maintaining the exact same parameter count (1.81 M) and computational complexity (2.28 GFLOPs) as the baseline model. This zero-overhead design contrasts sharply with other methods that require significantly more computational resources, making ERSR a highly practical and efficient solution for real-world clinical workflows.
 
\begin{table}[ht]
\centering
\caption{Analysis of model parameters and computational complexity.}
\label{numpara}
\renewcommand\arraystretch{0.9}
\setlength{\tabcolsep}{4.5mm}
\fontsize{8}{9}\selectfont
\begin{tabular}{c|c|c}
\specialrule{1pt}{0pt}{0pt} % 加粗横线，线宽 1pt
Models & Params(M) & GFLOPs \\ \hline
U-Net~\cite{ronneberger2015u} & 1.81      & 2.28   \\
BCP~\cite{bai2023bidirectional}    & 1.81      & 2.28   \\
DSTCT~\cite{jiang2024intrapartum}   & 27.15     & 5.91   \\
Mutual~\cite{su2024mutual} & 4.24      & 4.85   \\
SDCL~\cite{song2024sdcl}    & 1.81      & 2.28   \\
PICK~\cite{zeng2025pick}   & 3.27      & 2.86   \\
UniMatch V2~\cite{yang2025unimatch} & 24.18      & 7.57   \\
ERSR   & \textbf{1.81}      & \textbf{2.28}   \\ \specialrule{1pt}{0pt}{0pt} % 加粗横线，线宽 1pt
\end{tabular}
\end{table}

\subsection{Clinical significance of ERSR}
The proposed ERSR enhances the reliability and precision of automated fetal head segmentation. Its DS-AF and EC-PRe modules work in tandem to ensure robust segmentations. Furthermore, the SMCR strategy leverages the fetal head's intrinsic symmetry to ensure anatomically sound results. This directly improves the accuracy and reproducibility of critical biometric measurements (e.g., HC, BPD), enabling more precise tracking of fetal development and diagnosis of growth abnormalities. ERSR is a lightweight training strategy that requires minimal labeled data and adds no computational overhead during inference. Its accuracy, robustness, and efficiency make it a practical tool to accelerate clinical workflows, reduce annotation costs, and promote the adoption of automated tools for improved prenatal care.

\section{Conclusion}
In this work, we propose a novel framework, ERSR, for semi-supervised medical image segmentation of fetal head ultrasound images. ERSR consists of the dual-scoring adaptive filtering strategy, the ellipse-constrained pseudo-label refinement, and the symmetry-based multiple consistency regularization. The dual-scoring adaptive filtering strategy employs boundary consistency and contour regularity criteria to evaluate and filter teacher model outputs. The ellipse-constrained pseudo-label refinement refines these filtered outputs by fitting least squares ellipses to get finer pseudo-labels. The symmetry-based multiple consistency regularization strategy enforces consistency at multiple levels to ensure that the model learns robust and stable shape representations. Extensive experiments and ablation studies demonstrate that ERSR outperforms existing state-of-the-art SSL methods.

\section*{References}
\bibliographystyle{IEEEtran}
\bibliography{mybibliography}

\begin{thebibliography}{10}
\providecommand{\url}[1]{#1}
\csname url@samestyle\endcsname
\providecommand{\newblock}{\relax}
\providecommand{\bibinfo}[2]{#2}
\providecommand{\BIBentrySTDinterwordspacing}{\spaceskip=0pt\relax}
\providecommand{\BIBentryALTinterwordstretchfactor}{4}
\providecommand{\BIBentryALTinterwordspacing}{\spaceskip=\fontdimen2\font plus
\BIBentryALTinterwordstretchfactor\fontdimen3\font minus \fontdimen4\font\relax}
\providecommand{\BIBforeignlanguage}[2]{{%
\expandafter\ifx\csname l@#1\endcsname\relax
\typeout{** WARNING: IEEEtran.bst: No hyphenation pattern has been}%
\typeout{** loaded for the language `#1'. Using the pattern for}%
\typeout{** the default language instead.}%
\else
\language=\csname l@#1\endcsname
\fi
#2}}
\providecommand{\BIBdecl}{\relax}
\BIBdecl

\bibitem{kaindl2010many}
A.~M. Kaindl, S.~Passemard, P.~Kumar, N.~Kraemer, L.~Issa, A.~Zwirner, B.~Gerard, A.~Verloes, S.~Mani, and P.~Gressens, ``Many roads lead to primary autosomal recessive microcephaly,'' \emph{Progress in Neurobiology}, vol.~90, no.~3, pp. 363--383, 2010.

\bibitem{torres2022review}
H.~R. Torres, P.~Morais, B.~Oliveira, C.~Birdir, M.~R{\"u}diger, J.~C. Fonseca, and J.~L. Vila{\c{c}}a, ``A review of image processing methods for fetal head and brain analysis in ultrasound images,'' \emph{Computer Methods and Programs in Biomedicine}, vol. 215, p. 106629, 2022.

\bibitem{xodo2023fetal}
S.~Xodo, L.~Celante, S.~Liviero, M.~Orsaria, L.~Mariuzzi, M.~De~Luca, G.~Damante, L.~Driul, A.~Cagnacci, A.~Ferino \emph{et~al.}, ``{Fetal growth at term and placental oxidative stress in a tissue micro-array model: A histological and immunohistochemistry study},'' \emph{Histochemistry and Cell Biology}, vol. 160, no.~4, pp. 293--306, 2023.

\bibitem{chen2024direction}
Z.~Chen, Z.~Ou, Y.~Lu, and J.~Bai, ``Direction-guided and multi-scale feature screening for fetal head--pubic symphysis segmentation and angle of progression calculation,'' \emph{Expert Systems with Applications}, vol. 245, p. 123096, 2024.

\bibitem{ansari2024advancements}
M.~Y. Ansari, I.~A. Changaai~Mangalote, P.~K. Meher, O.~Aboumarzouk, A.~Al-Ansari, O.~Halabi, and S.~P. Dakua, ``{Advancements in deep learning for B-Mode ultrasound segmentation: A comprehensive review},'' \emph{IEEE Transactions on Emerging Topics in Computational Intelligence}, vol.~8, no.~3, pp. 2126--2149, 2024.

\bibitem{mushtaq2025comprehensive}
G.~Mushtaq and K.~Veningston, ``A comprehensive review on fetal health surveillance using ultrasound imagery,'' \emph{Discover Applied Sciences}, vol.~7, no.~5, pp. 1--38, 2025.

\bibitem{yu2019uncertainty}
L.~Yu, S.~Wang, X.~Li, C.-W. Fu, and P.-A. Heng, ``{Uncertainty-aware self-ensembling model for semi-supervised 3D left atrium segmentation},'' in \emph{Medical Image Computing and Computer Assisted Intervention -- MICCAI 2019}.\hskip 1em plus 0.5em minus 0.4em\relax Springer International Publishing, 2019, pp. 605--613.

\bibitem{cheplygina2019not}
V.~Cheplygina, M.~De~Bruijne, and J.~P. Pluim, ``Not-so-supervised: a survey of semi-supervised, multi-instance, and transfer learning in medical image analysis,'' \emph{Medical Image Analysis}, vol.~54, pp. 280--296, 2019.

\bibitem{bai2023bidirectional}
Y.~Bai, D.~Chen, Q.~Li, W.~Shen, and Y.~Wang, ``Bidirectional copy-paste for semi-supervised medical image segmentation,'' in \emph{Proceedings of the IEEE/CVF Conference on Computer Vision and Pattern Recognition}, 2023, pp. 11\,514--11\,524.

\bibitem{10246276}
P.-N. Bui, D.-T. Le, J.~Bum, S.~Kim, S.~J. Song, and H.~Choo, ``Semi-supervised learning with fact-forcing for medical image segmentation,'' \emph{IEEE Access}, vol.~11, pp. 99\,413--99\,425, 2023.

\bibitem{su2024mutual}
J.~Su, Z.~Luo, S.~Lian, D.~Lin, and S.~Li, ``Mutual learning with reliable pseudo label for semi-supervised medical image segmentation,'' \emph{Medical Image Analysis}, vol.~94, p. 103111, 2024.

\bibitem{song2024sdcl}
B.~Song and Q.~Wang, ``{SDCL: Students discrepancy-informed correction learning for semi-supervised medical image segmentation},'' in \emph{{Medical Image Computing and Computer Assisted Intervention -- MICCAI 2024}}.\hskip 1em plus 0.5em minus 0.4em\relax Springer Nature Switzerland, 2024, pp. 567--577.

\bibitem{bashir2024consistency}
R.~M.~S. Bashir, T.~Qaiser, S.~E.~A. Raza, and N.~M. Rajpoot, ``Consistency regularisation in varying contexts and feature perturbations for semi-supervised semantic segmentation of histology images,'' \emph{Medical Image Analysis}, vol.~91, p. 102997, 2024.

\bibitem{WANG2025112767}
Q.~Wang, D.~Zhao, H.~Ma, X.~Yang, and B.~Liu, ``Integrated generation adversarial and semi-supervised network for corpus callosum and cavum septum pellucidum complex segmentation in fetal brain ultrasound via progressive training,'' \emph{Applied Soft Computing}, vol. 171, p. 112767, 2025.

\bibitem{zeng2025pick}
Q.~Zeng, Z.~Lu, Y.~Xie, and Y.~Xia, ``{PICK: Predict and mask for semi-supervised medical image segmentation},'' \emph{International Journal of Computer Vision}, vol. 133, no.~6, pp. 3296--3311, Jun 2025.

\bibitem{meanteacher}
A.~Tarvainen and H.~Valpola, ``{Mean teachers are better role models: Weight-averaged consistency targets improve semi-supervised deep learning results},'' in \emph{Proceedings of the 31st International Conference on Neural Information Processing Systems}.\hskip 1em plus 0.5em minus 0.4em\relax Curran Associates Inc., 2017, p. 1195–1204.

\bibitem{zhang2017deep}
Y.~Zhang, L.~Yang, J.~Chen, M.~Fredericksen, D.~P. Hughes, and D.~Z. Chen, ``Deep adversarial networks for biomedical image segmentation utilizing unannotated images,'' in \emph{International Conference on Medical Image Computing and Computer-Assisted Intervention}.\hskip 1em plus 0.5em minus 0.4em\relax Springer, 2017, pp. 408--416.

\bibitem{xu2022all}
Z.~Xu, Y.~Wang, D.~Lu, L.~Yu, J.~Yan, J.~Luo, K.~Ma, Y.~Zheng, and R.~K.-y. Tong, ``{All-around real label supervision: Cyclic prototype consistency learning for semi-supervised medical image segmentation},'' \emph{IEEE Journal of Biomedical and Health Informatics}, vol.~26, no.~7, pp. 3174--3184, 2022.

\bibitem{lu2024dual}
S.~Lu, Z.~Yan, W.~Chen, T.~Cheng, Z.~Zhang, and G.~Yang, ``Dual consistency regularization with subjective logic for semi-supervised medical image segmentation,'' \emph{Computers in Biology and Medicine}, vol. 170, p. 107991, 2024.

\bibitem{huo2021atso}
X.~Huo, L.~Xie, J.~He, Z.~Yang, W.~Zhou, H.~Li, and Q.~Tian, ``{ATSO: Asynchronous teacher-student optimization for semi-supervised image segmentation},'' in \emph{Proceedings of the IEEE/CVF Conference on Computer Vision and Pattern Recognition}, 2021, pp. 1235--1244.

\bibitem{cascante2021curriculum}
P.~Cascante-Bonilla, F.~Tan, Y.~Qi, and V.~Ordonez, ``{Curriculum Labeling: Revisiting pseudo-labeling for semi-supervised learning},'' in \emph{Proceedings of the AAAI Conference on Artificial Intelligence}, vol.~35, no.~8, 2021, pp. 6912--6920.

\bibitem{liu2022semi}
J.~Liu, C.~Desrosiers, and Y.~Zhou, ``Semi-supervised medical image segmentation using cross-model pseudo-supervision with shape awareness and local context constraints,'' in \emph{International Conference on Medical Image Computing and Computer-Assisted Intervention}.\hskip 1em plus 0.5em minus 0.4em\relax Springer, 2022, pp. 140--150.

\bibitem{xu2024expectation}
M.~Xu, Y.~Zhou, C.~Jin, M.~de~Groot, D.~C. Alexander, N.~P. Oxtoby, Y.~Hu, and J.~Jacob, ``Expectation maximisation pseudo labels,'' \emph{Medical Image Analysis}, vol.~94, p. 103125, 2024.

\bibitem{jiang2024intrapartum}
J.~Jiang, H.~Wang, J.~Bai, S.~Long, S.~Chen, V.~M. Campello, and K.~Lekadir, ``{Intrapartum ultrasound image segmentation of pubic symphysis and fetal head using dual student-teacher framework with CNN-ViT collaborative learning},'' in \emph{International Conference on Medical Image Computing and Computer-Assisted Intervention}.\hskip 1em plus 0.5em minus 0.4em\relax Springer, 2024, pp. 448--458.

\bibitem{sun2024lcamix}
D.~Sun, F.~Dornaika, and J.~Charafeddine, ``{LCAMix: Local-and-contour aware grid mixing based data augmentation for medical image segmentation},'' \emph{Information Fusion}, vol. 110, p. 102484, 2024.

\bibitem{montagnat2001review}
J.~Montagnat, H.~Delingette, and N.~Ayache, ``A review of deformable surfaces: topology, geometry and deformation,'' \emph{Image and Vision Computing}, vol.~19, no.~14, pp. 1023--1040, 2001.

\bibitem{ronneberger2015u}
O.~Ronneberger, P.~Fischer, and T.~Brox, ``{U-Net: Convolutional networks for biomedical image segmentation},'' in \emph{Medical Image Computing and Computer-Assisted Intervention -- MICCAI 2015}, N.~Navab, J.~Hornegger, W.~M. Wells, and A.~F. Frangi, Eds.\hskip 1em plus 0.5em minus 0.4em\relax Springer International Publishing, 2015, pp. 234--241.

\bibitem{zhou2018unet++}
Z.~Zhou, M.~M. Rahman~Siddiquee, N.~Tajbakhsh, and J.~Liang, ``{UNet++: A nested U-Net architecture for medical image segmentation},'' in \emph{Deep Learning in Medical Image Analysis and Multimodal Learning for Clinical Decision Support}.\hskip 1em plus 0.5em minus 0.4em\relax Cham: Springer International Publishing, 2018, pp. 3--11.

\bibitem{oktay2018attention}
O.~Oktay, J.~Schlemper, L.~L. Folgoc, M.~Lee, M.~Heinrich, K.~Misawa, K.~Mori, S.~McDonagh, N.~Y. Hammerla, B.~Kainz \emph{et~al.}, ``{Attention U-Net: Learning where to look for the pancreas},'' \emph{arXiv preprint arXiv:1804.03999}, 2018.

\bibitem{7974824}
C.~F. Baumgartner, K.~Kamnitsas, J.~Matthew, T.~P. Fletcher, S.~Smith, L.~M. Koch, B.~Kainz, and D.~Rueckert, ``{SonoNet: Real-time detection and localisation of fetal standard scan planes in freehand ultrasound},'' \emph{IEEE Transactions on Medical Imaging}, vol.~36, no.~11, pp. 2204--2215, 2017.

\bibitem{OU2024108501}
Z.~Ou, J.~Bai, Z.~Chen, Y.~Lu, H.~Wang, S.~Long, and G.~Chen, ``{RTSeg-net: A lightweight network for real-time segmentation of fetal head and pubic symphysis from intrapartum ultrasound images},'' \emph{Computers in Biology and Medicine}, vol. 175, p. 108501, 2024.

\bibitem{zhang2017mixup}
H.~Zhang, M.~Cisse, Y.~N. Dauphin, and D.~Lopez-Paz, ``{Mixup: Beyond empirical risk minimization},'' \emph{arXiv preprint arXiv:1710.09412}, 2017.

\bibitem{devries2017improved}
T.~DeVries and G.~W. Taylor, ``Improved regularization of convolutional neural networks with cutout,'' \emph{arXiv preprint arXiv:1708.04552}, 2017.

\bibitem{yun2019cutmix}
S.~Yun, D.~Han, S.~J. Oh, S.~Chun, J.~Choe, and Y.~Yoo, ``{CutMix: Regularization strategy to train strong classifiers with localizable features},'' in \emph{Proceedings of the IEEE/CVF International Conference on Computer Vision}, 2019, pp. 6023--6032.

\bibitem{lee2013pseudo}
D.-H. Lee \emph{et~al.}, ``{Pseudo-Label: The simple and efficient semi-supervised learning method for deep neural networks},'' in \emph{Workshop on Challenges in Representation Learning, ICML}, vol.~3, no.~2.\hskip 1em plus 0.5em minus 0.4em\relax Atlanta, 2013, p. 896.

\bibitem{sohn2020fixmatch}
K.~Sohn, D.~Berthelot, N.~Carlini, Z.~Zhang, H.~Zhang, C.~A. Raffel, E.~D. Cubuk, A.~Kurakin, and C.-L. Li, ``{FixMatch: Simplifying semi-supervised learning with consistency and confidence},'' \emph{Advances in Neural Information Processing Systems}, vol.~33, pp. 596--608, 2020.

\bibitem{WANG2022102447}
K.~Wang, B.~Zhan, C.~Zu, X.~Wu, J.~Zhou, L.~Zhou, and Y.~Wang, ``Semi-supervised medical image segmentation via a tripled-uncertainty guided mean teacher model with contrastive learning,'' \emph{Medical Image Analysis}, vol.~79, p. 102447, 2022.

\bibitem{zeng2023pefat}
Q.~Zeng, Y.~Xie, Z.~Lu, and Y.~Xia, ``{PEFAT: Boosting semi-supervised medical image classification via pseudo-loss estimation and feature adversarial training},'' in \emph{IEEE/CVF Conference on Computer Vision and Pattern Recognition (CVPR)}, 2023, pp. 15\,671--15\,680.

\bibitem{huang2024dual}
Z.~Huang, D.~Gai, W.~Min, Q.~Wang, and L.~Zhan, ``Dual-stream-based dense local features contrastive learning for semi-supervised medical image segmentation,'' \emph{Biomedical Signal Processing and Control}, vol.~88, p. 105636, 2024.

\bibitem{li2024diversity}
W.~Li, R.~Bian, W.~Zhao, W.~Xu, and H.~Yang, ``{Diversity matters: Cross-head mutual mean-teaching for semi-supervised medical image segmentation},'' \emph{Medical Image Analysis}, vol.~97, p. 103302, 2024.

\bibitem{zeng2024reciprocal}
Q.~Zeng, Z.~Lu, Y.~Xie, M.~Lu, X.~Ma, and Y.~Xia, ``Reciprocal collaboration for semi-supervised medical image classification,'' in \emph{Medical Image Computing and Computer Assisted Intervention -- MICCAI 2024}, M.~G. Linguraru, Q.~Dou, A.~Feragen, S.~Giannarou, B.~Glocker, K.~Lekadir, and J.~A. Schnabel, Eds.\hskip 1em plus 0.5em minus 0.4em\relax Cham: Springer Nature Switzerland, 2024, pp. 522--532.

\bibitem{zeng2025uncertainty}
X.~Zeng, S.~Xiong, J.~Xu, G.~Du, and Y.~Rong, ``Uncertainty co-estimator for improving semi-supervised medical image segmentation,'' \emph{IEEE Transactions on Medical Imaging}, 2025.

\bibitem{zeng2025segment}
Q.~Zeng, Y.~Xie, Z.~Lu, M.~Lu, Y.~Wu, and Y.~Xia, ``{Segment Together: A versatile paradigm for semi-supervised medical image segmentation},'' \emph{IEEE Transactions on Medical Imaging}, vol.~44, no.~7, pp. 2948--2959, 2025.

\bibitem{zeng2025consistency}
Q.~Zeng, Y.~Xie, Z.~Lu, M.~Lu, J.~Zhang, and Y.~Xia, ``Consistency-guided differential decoding for enhancing semi-supervised medical image segmentation,'' \emph{IEEE Transactions on Medical Imaging}, vol.~44, no.~1, pp. 44--56, 2025.

\bibitem{tang2025swma-unet}
X.~Tang, J.~Li, Q.~Liu, C.~Zhou, P.~Zeng, Y.~Meng, J.~Xu, G.~Tian, and J.~Yang, ``{SWMA-UNet: Multi-path attention network for improved medical image segmentation},'' \emph{IEEE Journal of Biomedical and Health Informatics}, vol.~29, no.~5, pp. 3609--3618, 2025.

\bibitem{laine2016temporal}
S.~Laine and T.~Aila, ``Temporal ensembling for semi-supervised learning,'' \emph{arXiv preprint arXiv:1610.02242}, 2016.

\bibitem{van2018automated}
T.~L. van~den Heuvel, D.~de~Bruijn, C.~L. de~Korte, and B.~v. Ginneken, ``{Automated measurement of fetal head circumference using 2D ultrasound images},'' \emph{PloS One}, vol.~13, no.~8, p. e0200412, 2018.

\bibitem{lu2022jnu}
Y.~Lu, M.~Zhou, D.~Zhi, M.~Zhou, X.~Jiang, R.~Qiu, Z.~Ou, H.~Wang, D.~Qiu, M.~Zhong \emph{et~al.}, ``{The JNU-IFM dataset for segmenting pubic symphysis-fetal head},'' \emph{Data in Brief}, vol.~41, p. 107904, 2022.

\bibitem{wu2021r}
X.~Liang, L.~Wu, J.~Li, Y.~Wang, Q.~Meng, T.~Qin, W.~Chen, M.~Zhang, T.-Y. Liu \emph{et~al.}, ``{R-Drop: Regularized dropout for neural networks},'' \emph{Advances in Neural Information Processing Systems}, vol.~34, pp. 10\,890--10\,905, 2021.

\bibitem{yang2025unimatch}
L.~Yang, Z.~Zhao, and H.~Zhao, ``{Unimatch v2: Pushing the limit of semi-supervised semantic segmentation},'' \emph{IEEE Transactions on Pattern Analysis and Machine Intelligence}, 2025.

\bibitem{chen2025comt}
Y.~Chen, B.~Deng, Z.~Peng, Y.~Zhou, P.~Sun, J.~Wan, and J.~Bai, ``{COMT: Co-training mean teachers semi-supervised training framework for cervical segmentation},'' in \emph{2025 IEEE 22nd International Symposium on Biomedical Imaging (ISBI)}.\hskip 1em plus 0.5em minus 0.4em\relax IEEE, 2025, pp. 1--5.

\bibitem{yu2025enhancing}
J.~Yu, ``Enhancing cervical segmentation via alternate diverse teaching in semi-supervised learning,'' in \emph{2025 IEEE 22nd International Symposium on Biomedical Imaging (ISBI)}.\hskip 1em plus 0.5em minus 0.4em\relax IEEE, 2025, pp. 1--4.

\bibitem{xiao2025hierarchical}
Y.~Xiao and L.~Xiao, ``Hierarchical augmentation consistency learning for semi-supervised medical image segmentation,'' in \emph{2025 IEEE 22nd International Symposium on Biomedical Imaging (ISBI)}.\hskip 1em plus 0.5em minus 0.4em\relax IEEE, 2025, pp. 1--5.

\bibitem{jiang2025semi}
J.~Jiang, Y.~Bi, C.~Zhou, Y.~Liu, and J.~Zhang, ``{Semi-Cervixseg: A multi-stage training strategy for semi-supervised cervical segmentation},'' in \emph{2025 IEEE 22nd International Symposium on Biomedical Imaging (ISBI)}.\hskip 1em plus 0.5em minus 0.4em\relax IEEE, 2025, pp. 1--5.

\bibitem{bai2025psfhs}
J.~Bai, Z.~Zhou, Z.~Ou, G.~Koehler, R.~Stock, K.~Maier-Hein, M.~Elbatel, R.~Mart{\'\i}, X.~Li, Y.~Qiu \emph{et~al.}, ``{PSFHS challenge report: pubic symphysis and fetal head segmentation from intrapartum ultrasound images},'' \emph{Medical Image Analysis}, vol.~99, p. 103353, 2025.

\bibitem{sappia2025acouslic}
M.~S. Sappia, C.~L. de~Korte, B.~van Ginneken, D.~Ninalga, S.~Kondo, S.~Kasai, K.~Hirasawa, T.~Akumu, C.~Mart{\'\i}n-Isla, K.~Lekadir \emph{et~al.}, ``{ACOUSLIC-AI challenge report: Fetal abdominal circumference measurement on blind-sweep ultrasound data from low-income countries},'' \emph{Medical Image Analysis}, p. 103640, 2025.

\end{thebibliography}
\end{document}